# Benchmarking Lane-changing Decision-making for Deep Reinforcement Learning

JUNJIE WANG[1,2], QICHAO ZHANG[1,2], DONGBIN ZHAO[1,2]

1. The State Key Laboratory of Management and Control for Complex Systems, Institute of Automation, Chinese Academy of Sciences, Beijing, China;

2. College of Artificial Intelligence, University of Chinese Academy of Sciences, Beijing, China

The development of autonomous driving has attracted extensive attention in recent years, and it is essential to evaluate the performance of autonomous driving. However, testing on the road is expensive and inefficient. Virtual testing is the primary way to validate and verify self-driving cars, and the basis of virtual testing is to build simulation scenarios. In this paper, we propose a training, testing, and evaluation pipeline for the lane-changing task from the perspective of deep reinforcement learning. First, we design lane change scenarios for training and testing, where the test scenarios include stochastic and deterministic parts. Then, we deploy a set of benchmarks consisting of learning and non-learning approaches. We train several state-of-the-art deep reinforcement learning methods in the designed training scenarios and provide the benchmark metrics evaluation results of the trained models in the test scenarios. The designed lane-changing scenarios and benchmarks are both opened to provide a consistent experimental environment for the lane-changing task[1].

## 1 INTRODUCTION

In recent years, autonomous driving technology has been developing rapidly. It is expected that by 2050, the application of this technology can reduce vehicle emissions by 50%, and the road traffic casualty rate will be close to zero [1]. For industry players, the main testing method is the real vehicle road test. However, Kalra et al. [2] of RAND Corporation conclude that at the 95% confidence level, road testing of more than 14.2 billion km is required to prove that the fatality rate of autonomous vehicles is 20% lower than that of human drivers. Therefore, virtual testing will be the primary way of validation and verification of autonomous vehicles.

Reinforcement Learning (RL) agents learn by interacting with the environment, adjust their policy by obtaining rewards, and maximize the reward function by balancing exploration and exploitation, expecting to find the optimal policy corresponding to the maximum cumulative reward [3]. Deep Reinforcement Learning (DRL), combining the perception capability of Deep Learning (DL) and the decision-making capability of RL [4], is suitable for solving the autonomous driving decision-making problem, which is a typical application of time-series decisions in a complex environment. Many existing studies apply DRL to the intersection [5], lane changing [6], [7] scenarios, etc. Still, to the best of our knowledge, there is no standardized system for training and testing scenarios, evaluation metrics, and baseline methods performance comparisons.

---

[1] The code of our proposed experimental environment can be found at https://github.com/powerfulwang/CarlaHighwayTest.

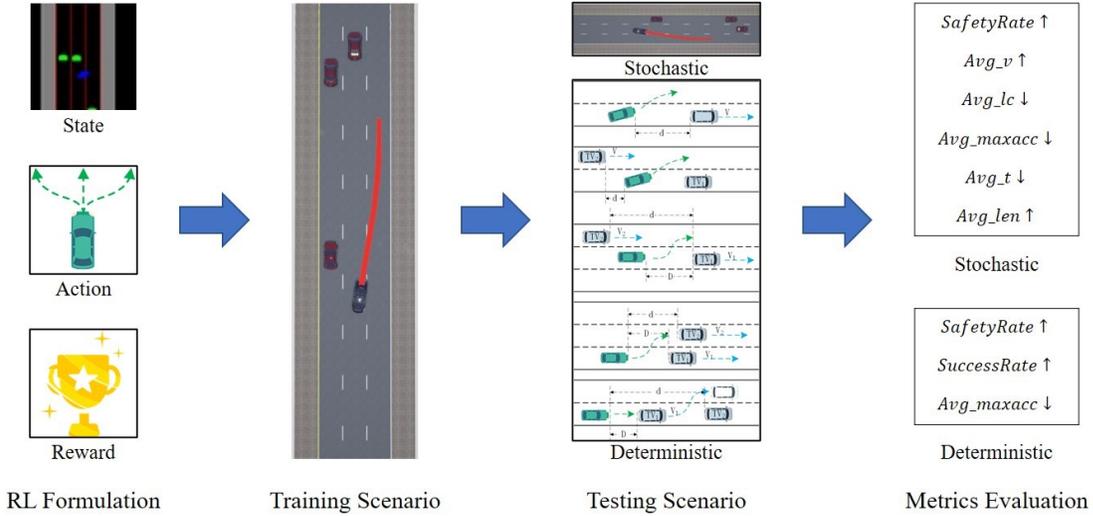

Figure 1: The proposed pipeline of training, testing, and evaluation.

In this paper, we propose a training, testing, and evaluation pipeline (shown in Figure 1) for lane change scenarios from the perspective of DRL with the help of a virtual autonomous driving simulator. The contributions of our work are listed as follows: i) We design training and testing scenarios for the lane-changing task, where the test scenarios include stochastic and deterministic parts (Section 2); ii) We provide evaluation metrics for the agents in the stochastic and deterministic test scenarios, respectively (Section 5); iii) We formalize the lane-changing decision-making problem as an image-state-based markov decision process with definitions of actions and rewards (Section 3), and provide benchmark results for several algorithms, including the non-learning method MOBIL and state-of-the-art DRL methods D3QN, A2C, and PPO under our design (Section 4 and 6).

## 2 LANE CHANGE SCENARIOS

The lane changing scenario is a typical class of driving scenarios involving multiple traffic participants. This paper first introduces the conducted training and testing scenarios of lane changing.

We use the CARLA [8] simulator, which consists of a scalable client-server architecture. The server is responsible for the simulation itself: sensor rendering, physics calculations, updates of the world state and its participants, etc. The client consists of a series of modules that control the logic and set the world conditions of actors (an actor is any entity in the simulation world, such as a vehicle, sensor, traffic light, etc.) in the scenario.

### 2.1 Training Scenario

The dynamic training scenario is constructed with random traffic flows, which are realized with the help of the traffic manager tool in CARLA simulator version 0.9.9. We first build a map of a three-lane highway that can be imported into CARLA (also open-source with our code) and select a part for training scenario generation. For training, the vehicle controlled by the algorithm is called the ego vehicle, and other vehicles are called the social vehicles.

The random training scenario generation method is as follows: firstly, the ego vehicle is randomly generated at the lane center of the road section (three lanes are random), and then 6 to 12 social vehicles are randomly generated at the lane center within the range of 30 m behind and 180 m in front of the ego vehicle. The target speed of the ego vehicle is 60 km/h, and the expected speed of social vehicles is a random value from 20 to 40 km/h. The following distance between one social vehicle and its front vehicle is random from 0 to 15 m. Social vehicles will have unexpected lane-changing behavior (controlled by the traffic manager). And the lane width is 3.5 m. The screenshot of the training scenario is illustrated in Figure 2. Since the number, position, and speed of surrounding social vehicles in the training scenario are random, and there will be unexpected lane change behaviors, a series of challenging instances can be generalized. Note that the training scenario does not contain low-speed (speed less than 20km/h) social vehicles, and the processing of the low-speed scenario can reflect the generalization of an agent to a certain extent.

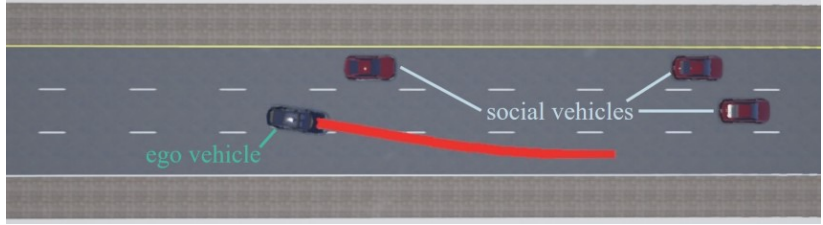

Figure 2: The training scenario.

## 2.2 Testing Scenarios

### 2.2.1 Stochastic Test.

For the stochastic test scenario, its setting is similar to the training scenario. For the test scenario, if the surrounding vehicle density is high, the ego vehicle can hardly change lanes. Such a test scenario is of little significance. Therefore, 4 to 9 social vehicles will be randomly generated around the ego vehicle. At the same time, social vehicles are restricted from random lane changes and will stay in their initial lane.

### 2.2.2 Deterministic Test.

We refer to the group standard [9] for our deterministic test scenarios design. We design five major logic scenarios of lane-changing tests and obtain more than 400 specific test scenarios through parameter instantiation. For deterministic testing, the vehicle controlled by algorithms for lane changing is called the test vehicle, and the other environmental vehicles are called the target vehicles. The lane-changing model trained in the training scenario is tested in the test scenarios with statistical metrics.

The demonstration of the designed logical scenarios of lane changing is shown in Figure 3. The Test Vehicle (TV) travels on a straight road with target speed $V_T$. Dashed lines indicate the lanes that can be changed to, while solid lines indicate the lanes that cannot be changed to, and the lane's width is $X_0$. $d$ and $D$ in Figure 3 denote the distance between TV and the target vehicle $GV$. Here we take the subfigure (c) as an example to explain the parameters in Figure 3. For (c), a target vehicle $GV_1$ traveling at speed $V_1$ along the centerline of the lane is placed in front of TV, and the longitudinal distance between TV and $GV_1$ is $D$. The adjacent lane on the left side of TV places a target vehicle $GV_2$ traveling at speed $V_2$ along the centerline of the lane, and the longitudinal distance between TV and $GV_2$ is $d$. These test scenarios provide a range of different examples to test the intelligence of the agent of TV. For example, (c) is a three-lane scenario where there has a vehicle present in the left lane and no vehicles in the right lane, and the TV needs to decide which way to change lanes. (e) is a two-lane scenario where the TV needs to react in time to the sudden appearance of a stationary vehicle.

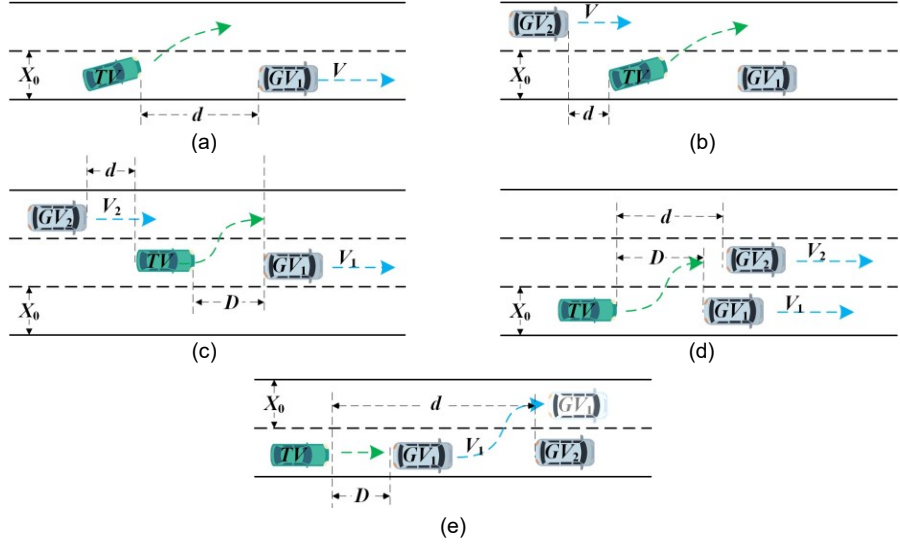

Figure 3: The deterministic testing scenarios.

## 3 REINFORCEMENT LEARNING FORMULATION

A bird's eye view image of the perceived range of the ego vehicle is used as input to the algorithm, and the output of the algorithm is a high-level lane change command. The task of image-based lane change decision-making is modeled as a Markov Decision Process (MDP). An MDP is a tuple of the form: $\langle S, A, P, R \rangle$, where $S$ is a finite set of states, $A$ a finite set of actions, $P$ the dynamic transition model $P(s_{t+1} = s'|s_t = s, a_t = a)$ for each action, $R$ the reward function $R(s_t = s, a_t = a) = \mathbb{E}(r_t|s_t = s, a_t = a)$, and $\gamma \in [0,1]$ the discount factor. The state transition model $P$ and reward $R$ are affected by the specific behavior $a$. The goal of RL is to learn a policy $\pi^*(a|s)$ that maximizes the cumulative reward $J_\pi = \mathrm{argmax}_\pi \mathbb{E} \sum_t \gamma^{t-1} r_t$. Consider an environment $E$ in which the agent interacts with it through a set of states, actions, and rewards. In this paper, the state, action space, and reward function are set as follows.

### 3.1 State

The state input is a three-channel RGB image of size 64×64, corresponding to the overhead traffic flow within 50 m in front of the ego vehicle and 25 m behind it. The ego vehicle is blue; the social vehicles are green; the lane line is red; and there is a one-lane-wide sidewalk on each side of the road, filled with gray. The position of the ego vehicle in the image is relatively fixed.

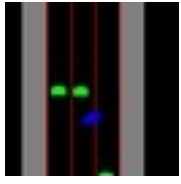

Figure 4: The state of scenarios.

### 3.2 Action Space

The output of DRL algorithms is a lateral lane change command, corresponding to a discrete action space set as {keep current lane, change lanes to the left, and change lanes to the right}.

### 3.3 Reward Design

The reward function integrates the safety and efficiency of the decision action. The efficiency is reflected in the speed of the ego vehicle on the one hand. To be specific, a positive reward $r_v = 0.2 \cdot v/v_T$ is defined, where $v$ is the ego speed and $v_T = 60$ km/h is the desired speed of the ego vehicle. On the other hand, a negative reward of $r_l = -1.0$ is given to the agent when a lane change occurs, at which time the speed reward will no longer be applied. Safety is mainly reflected in the fact that no collision can occur, and if a collision between the ego vehicle and other vehicles happens within a decision-making period, a penalty of $-1.0$ will be applied on the basis of the reward mentioned above, and the collision reward is defined as $r_c$. In summary, the total reward function is defined as

$$r_c = \begin{cases} 0 & \text{no collision} \\ -1.0 & \text{a collision happens} \end{cases}, \quad (1)$$

$$r = \begin{cases} r_v + r_c & \text{lane keeping} \\ r_l + r_c & \text{lane changing} \end{cases}. \quad (2)$$

After formalizing the problem as an MDP, we can utilize existing DRL methods to train the agent with the ability to make autonomous lane change decisions.

## 4 BASELINE METHODS

We choose the non-learning method MOBIL (Minimizing Overall Braking Induced by Lane Changes) [7] and several state-of-the-art DRL algorithms D3QN (Dueling Double DQN) [14], [15], [16], PPO (Proximal Policy Optimization) [11], and A2C (Advantage Actor-Critic) [12] as our baselines.

### 4.1 MOBIL

MOBIL is a lane change model of vehicle following, which considers the difference of vehicle acceleration (or deceleration) after lane change as the benefit value, and applies braking deceleration to the new follower in the target lane to avoid accidents. For the acceleration $\tilde{a}_e$ of the ego vehicle after lane change and the acceleration $\tilde{a}_n$ of the new following vehicle on the target lane after ego lane changing, $\tilde{a}_e > -b_{safe}$ and $\tilde{a}_n > -b_{safe}$ should be satisfied, where $b_{safe}$ is the maximum safe deceleration. The IDM model [13] is used to predict the acceleration of the ego and surrounding vehicles. If the safety criterion is met, MOBIL changes lanes if

$$\tilde{a}_e - a_e + p((\tilde{a}_n - a_n) + (\tilde{a}_o - a_o)) > a_{th}, \quad (3)$$

where $a_e$, $a_n$, and $a_o$ are the current accelerations of the target vehicle, the vehicle following the target lane, and the vehicle following the original lane, respectively. The wavy line symbol ˜ indicates the corresponding accelerations after the lane change is implemented. $\rho$ is the courtesy factor, and $a_{th}$ is the lane change benefit threshold. We set the priority to left when both left and right lane changes are allowed.

Note that, unlike RL algorithms, the non-learning method MOBIL uses not image-based state inputs but the true values of the desired states.

### 4.2 D3QN

D3QN is Deep Q-Network (DQN) [14] combined with dueling [15], double [16] framework. DQN is a method that consolidates neural networks and Q learning.

The weights in the neural network are denoted by $\theta$, and the action-value function fitted by the neural network is denoted by $Q(s, a; \theta)$ with the target network denoted by $Q'(s, a; \theta')$. Dueling DQN considers dividing the Q network into two parts. The first part is only related to the state $s$ and not to the specific action $a$ to be adopted. This part is called the value function part, which is written as $V(s, \theta, \alpha)$. The second part is related to both the state $s$ and the action $a$ and is called the advantage function part, denoted as $A(s, a, \theta, \beta)$. Then finally the value function can be re-expressed as $Q(s, a, \theta, \alpha, \beta) = V(s, \theta, \alpha) + A(s, a, \theta, \beta)$. Considering that the actual Q function value is approximated by the neural network as much as possible, the loss function of the neural network training can be defined as

$$L_i(\theta_i) = \mathbb{E}_{s \sim \pi}\left[\frac{1}{2}(y_i - Q(s, a; \theta_i))^2\right], \quad (4)$$

where $y_i = \mathbb{E}_{s' \sim E}[r + \gamma Q'(s', \mathrm{argmax}_{a'} Q(s', a; \theta_i); \theta'_i)|s, a]$ is the objective value of the $i$-th iteration in Double DQN. The above loss function for the gradient of the weights yields

$$\nabla_{\theta_i} L_i(\theta_i) = \mathbb{E}_{s' \sim E}[r + \gamma Q'(s', \mathrm{argmax}_{a'} Q(s', a; \theta_i); \theta'_i) - Q(s, a; \theta_i))\nabla_{\theta_i} Q(s, a; \theta_i)]. \quad (5)$$

For computational convenience, the stochastic gradient descent method is usually used to optimize the loss function instead of directly computing the full expectation value in the above equation.

### 4.3 A2C

In synchronous AC, all of the updates by the parallel agents are collected to update the global network. To encourage exploration, stochastic noise is added to the probability distribution of the actions predicted by each agent. Let $A(s, a) = Q(s, a) - V(s)$ be the advantage function (the advantage of action $a$ relative to the average performance), the gradient of the A2C algorithm can be obtained by

$$\nabla_\theta J(\theta) = \mathbb{E}_{\pi_\theta}[\nabla_\theta \log \pi_\theta(s, a) A(s, a)]. \quad (6)$$

In practice, it is not necessary to maintain two sets of parameters to interactively approximate $Q(s, a)$ and $V(s)$, respectively. Specifically, we can use $\delta^A = r + \lambda V(s') - V(s)$ instead of $\delta = r + \lambda Q(s', a') - Q(s, a)$, since by definition $\mathbb{E}(\delta) = \delta^A$. And it happens that $\delta^A$ is the unbiased estimate of $A(s, a)$, because, by definition of the Q function, there is $\mathbb{E}[r + \lambda V(s')|s, a] = Q(s, a)$. Therefore, in fact, when implementing A2C algorithm, only one set of parameters need to be maintained for estimating $V(s)$, and doing gradient descent to update the parameters can be done using

$$\Delta \theta = \alpha \nabla_\theta \log \pi_\theta(s, a)(r + \lambda V(s') - V(s)). \quad (7)$$

### 4.4 PPO

PPO is a Policy Gradient (PG) algorithm. The PG algorithms are susceptible to the learning step size, and it is challenging to select an appropriate one. The difference between the old and new policies during training is not conducive to learning if it is too large. PPO proposes a new objective function that can be updated in small batches at multiple training steps, solving the problem of difficult to determine the learning step size in PG algorithms. The clip variant of PPO eliminates the incentive between the new policy and the old policy by clipping the objective function specifically. It updates policy $\pi_\theta$ with

$$\theta_{k+1} = \underset{\theta}{\operatorname{argmax}} \mathbb{E}_{s,a \sim \pi_{\theta_k}}[L(s, a, \theta_k, \theta)]. \quad (8)$$

And $L(s, a, \theta_k, \theta)$ is given by

$$L(s, a, \theta_k, \theta) = \min\left(\frac{\pi_\theta(a|s)}{\pi_{\theta_k}(a|s)} A^{\pi_{\theta_k}}(s, a), \operatorname{clip}\left(\frac{\pi_\theta(a|s)}{\pi_{\theta_k}(a|s)}, 1 - \epsilon, 1 + \epsilon\right) A^{\pi_{\theta_k}}(s, a)\right), \quad (9)$$

where $\epsilon$ is a small hyperparameter that controls the distance between the new policy and the old one, and also $A^\pi(s, a) = Q^\pi(s, a) - V^\pi(s)$ is the advantage of an action $a$.

## 5 PERFORMANCE EVALUATION METRICS

Many metrics can be used to measure the behavior of agents [17]. For an autonomous driving system, safety and efficiency are the most concerned performance metrics.

For stochastic test scenarios, the defined metrics including $SafetyRate$, that is, the percentage of all episodes that do not collide, and $Avg\_v$, $Avg\_lc$, $Avg\_maxacc$, $Avg\_t$, and $Avg\_len$, the average speed, average lane change times, average maximum acceleration, average episode time of success runs, and average passage distance for all test episodes, respectively.

For deterministic test scenarios, we use lane change safety rate $SafetyRate$, lane change success rate $SuccessRate$, and average maximum acceleration $Avg\_maxacc$ as metrics to evaluate the performance of the test agents. Lane change safety rate is the percentage of safe lane change scenarios $SafeCount$ over test instances for all scenarios $TotalCount$:

$$SafetyRate = \frac{SafeCount}{TotalCount} \times 100\%. \quad (10)$$

Lane change success rate is the percentage of successful lane change scenarios $SuccessCount$ over the number of test instances for all scenarios $TotalCount$:

$$SuccessRate = \frac{SuccessCount}{TotalCount} \times 100\%. \quad (11)$$

And average maximum acceleration is the average of the vehicle's maximum acceleration under test for all scenarios. We define the scenario in which the lane change behavior causes no collision during the scenario test as the safe lane change scenario and the scenario in which the main vehicle is in a lane other than its initial lane at the end of the scenario as the success lane change scenario.

## 6 SIMULATION EXPERIMENTS

**Training.** We first train the DRL agents in the training scenario described in Section 2.1, where the RL problem is formalized in the formulation introduced in Section 3. The learning curves of RL baselines in the task of training scenario are depicted in Figure 5. The figure shows that A2C and PPO have faster learning efficiency than D3QN, but all three have similar final learning performance in the training scenario.

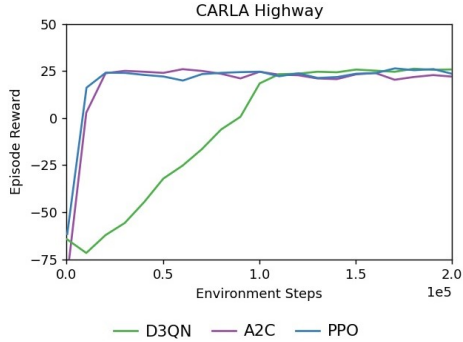

Figure 5: The learning curves of RL baselines in the task of training scenario.

**Testing.** We further test the trained agents in the designed stochastic and deterministic scenarios. Results are summarized in Tables 1 to 3. Table 1 shows the test results of different agents in the stochastic test scenario. Tables 2 and 3 give the test results of a single class logic scenario under the deterministic test scenarios and the statistical results under all scenarios, respectively. Some basic rules are added during the test to constrain the agents' behavior, similar to the work done in [18]. Note that the MOBIL method uses truth information instead of image state. It can be seen from Table 1 that in the stochastic test scenario, the A2C agent can hardly change lanes, while D3QN and PPO agents have a certain lane-changing ability. D3QN achieves the best safety among several methods, but its efficiency is lower than MOBIL. It is worth noting that although the PPO method has slightly more lane change times, it exceeds MOBIL in terms of all other metrics, showing the capability of DRL-based methods.

Table 1: The statistical results of stochastic test.

| Agent | MOBIL[2] | D3QN | A2C | PPO |
|---|---|---|---|---|
| $SafetyRate \uparrow$ | 96.0% | **97.6%** | 94.4% | **97.2%** |
| $Avg\_v \uparrow$ | 41.96 | 38.13 | 28.52 | **42.18** |
| $Avg\_lc \downarrow$ | 1.45 | 2.07 | **0.02** | 2.15 |
| $Avg\_maxacc \downarrow$ | **2.37** | 2.46 | 2.60 | **2.36** |
| $Avg\_t \downarrow$ | 35.34 | 41.54 | 62.37 | **34.71** |
| $Avg\_len \uparrow$ | 388.72 | 388.36 | 381.29 | **390.24** |

As can be seen from Table 2 and Table 3, the PPO agent still performs relatively well in deterministic test scenarios, and it achieves comparable performance to MOBIL. Taking all the test results together, the PPO method performs best under our setting, while the non-learning method, MOBIL, provides a strong baseline. The A2C method does not learn good lane change behavior, while D3QN performs relatively mediocre.

---

[2] The input of MOBIL is the vector representation with true state values, not the image-based state representation used in DRL methods.

Table 2: Test results of every deterministic logical scenario class.

| Logical scenario class | | (a) | (b) | (c) | (d) | (e) |
|---|---|---|---|---|---|---|
| Total scenario count | | 10 | 30 | 231 | 126 | 25 |
| Collision count | MOBIL[2] | 0 | 0 | 0 | 0 | 2 |
| | D3QN | 4 | 30 | 0 | 2 | 4 |
| | A2C | 5 | 30 | 0 | 0 | 25 |
| | PPO | 0 | 0 | 0 | 1 | 2 |
| Failure count | MOBIL[2] | 0 | 0 | 1 | 14 | 0 |
| | D3QN | 0 | 0 | 0 | 10 | 8 |
| | A2C | 6 | 0 | 231 | 126 | 0 |
| | PPO | 0 | 0 | 2 | 4 | 6 |
| Average maximum acceleration | MOBIL[2] | 0.40 | 0.41 | 0.40 | 1.33 | 0.79 |
| | D3QN | 1.09 | 2.05 | 0.71 | 1.64 | 1.33 |
| | A2C | 2.87 | 2.04 | 3.23 | 3.23 | 2.40 |
| | PPO | 0.39 | 0.39 | 0.53 | 1.21 | 1.47 |

Table 3: The statistical results of deterministic test scenarios.

| Agent | MOBIL[2] | D3QN | A2C | PPO |
|---|---|---|---|---|
| $SafetyRate \uparrow$ | **99.5%** | 90.5% | 85.8% | **99.3%** |
| $SuccessRate \uparrow$ | 96.4% | 95.7% | 14.0% | **97.2%** |
| $Avg\_maxacc \downarrow$ | **0.70** | 1.13 | 3.09 | 0.77 |

## 7 CONCLUSION

In this paper, we propose a training, testing, and evaluation pipeline for lane change scenarios. Under our formalization of the reinforcement learning problem, we construct training scenarios to train several state-of-the-art reinforcement learning agents, namely, D3QN, A2C, and PPO. Also, we implement the non-learning lane-changing method MOBIL as a comparison baseline. Furthermore, we construct test scenarios to evaluate the learned models and provide benchmark results under the evaluation metrics. The designed experimental environment is opened to promote the research development of the lane-changing task.

## ACKNOWLEDGMENTS

This work was supported in part by the National Natural Science Foundation of China (NSFC) under Grants No. 61803371, the Beijing Science and Technology Plan under Grants Z191100007419002, and the Beijing Municipal Natural Science Foundation under Grants L191002.